\documentclass{article}

\usepackage{arxiv}

\usepackage[utf8]{inputenc} % allow utf-8 input
\usepackage[T1]{fontenc}    % use 8-bit T1 fonts
\usepackage{hyperref}       % hyperlinks
\usepackage{url}            % simple URL typesetting
\usepackage{booktabs}       % professional-quality tables
\usepackage{amsfonts}       % blackboard math symbols
\usepackage{nicefrac}       % compact symbols for 1/2, etc.
\usepackage{microtype}      % microtypography
\usepackage{amsmath}        % add math by myselt
\usepackage{cleveref}       % smart cross-referencing
\usepackage{lipsum}         % Can be removed after putting your text content
\usepackage{graphicx}
\usepackage{natbib}
\usepackage{doi}
\usepackage{caption}       % add by myselt

\title{Solving Nonlinear Energy Supply and Demand System Using  Physics-Informed Neural Networks}

\newif\ifuniqueAffiliation
\ifuniqueAffiliation % Standard variant of author block
\author{ \href{https://orcid.org/0009-0008-2701-4775}{\includegraphics[scale=0.06]{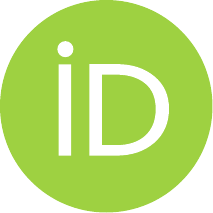}\hspace{1mm}Van Truong Vo}\\
	Scientific Research Department\\
	Irkutsk National Research Technical University\\
	664074 Irkutsk, Russia \\
	\texttt{vvtruong@ictu.edu.vn} \\
	\And
	\href{https://orcid.org/0000-0002-2307-0891}{\includegraphics[scale=0.06]{orcid.pdf}\hspace{1mm}Samad Noeiaghdam } \\
	Institute of Mathematics\\
	Henan Academy of Sciences\\
	Zhengzhou, 450046, China \\
	\texttt{snoei@hnas.ac.cn} \\
	\And
	\href{https://orcid.org/0000-0002-3131-1325}{\includegraphics[scale=0.06]{orcid.pdf}\hspace{1mm}Denis Sidorov } \\
	Sino--Russian Joint Research Center for Advanced Energy \& Power Systems\\
	Melentiev Energy Systems Institute\\
	Siberian Branch of Russian Academy of Sciences \\
	Irkutsk 664003, Russia\\
	\texttt{dsidorov@isem.irk.ru} \\
	\And
	\href{https://orcid.org/0000-0002-5032-0665}{\includegraphics[scale=0.06]{orcid.pdf}\hspace{1mm}Aliona Dreglea } \\
	Scientific Research Department\\
	Irkutsk National Research Technical University\\
	664074 Irkutsk, Russia\\
	\texttt{adreglea@gmail.com} \\
	\AND
	Liguo Wang \\
	School of Electrical Engineering and Automation \\
	Harbin Institute of Technology\\
	Harbin, China \\
	\texttt{wlg2001@hit.edu.cn} \\
}
\else
\usepackage{authblk}

\newbox{\orcid}\sbox{\orcid}{\includegraphics[scale=0.06]{orcid.pdf}} 
\author[1]{%
	\href{https://orcid.org/0009-0008-2701-4775}{\usebox{\orcid}\hspace{1mm}Van Truong Vo\thanks{\texttt{vvtruong@ictu.edu.vn}}}%
}
\author[2]{%
	\href{https://orcid.org/0000-0002-2307-0891}{\usebox{\orcid}\hspace{1mm}Samad Noeiaghdam\thanks{\texttt{snoei@hnas.ac.cn}}}%
}
\author[1,3,4]{%
	\href{https://orcid.org/0000-0002-3131-1325}{\usebox{\orcid}\hspace{1mm}Denis Sidorov\thanks{\texttt{dsidorov@isem.irk.ru}}}%
}
\author[1,4]{%
	\href{https://orcid.org/0000-0002-5032-0665}{\usebox{\orcid}\hspace{1mm}Aliona Dreglea\thanks{\texttt{adreglea@gmail.com}}}%
}
\author[4]{%
	\href{}{\hspace{1mm}Liguo Wang\thanks{\texttt{wlg2001@hit.edu.cn}}}%
}
\affil[1]{Scientific Research Department, Irkutsk National Research Technical University, 664074 Irkutsk, Russia}
\affil[2]{Institute of Mathematics, Henan Academy of Sciences, Zhengzhou, 450046, China}
\affil[3]{Applied Mathematics Department, Melentiev Energy Systems Institute, Siberian Branch of Russian Academy of Sciences, Irkutsk 664003, Russia}
\affil[4]{School of Electrical Engineering and Automation, Harbin Institute of Technology, Harbin, China}
\fi

\renewcommand{\shorttitle}{Solving Nonlinear Energy Supply and Demand System Using  PINNs}

\hypersetup{
pdftitle={Solving Nonlinear Energy Supply and Demand System Using  Physics-Informed Neural Networks},
pdfsubject={cs.AI, cs.LG},
pdfauthor={Van Truong Vo, Samad Noeiaghdam, Denis Sidorov, Aliona Dreglea, Liguo Wang},
pdfkeywords={Nonlinear energy supply and demand system, Physics-Informed Neural Networks, Machine learning, Deep learning, Numerical method.},
}

\begin{document}
\maketitle

\begin{abstract}
	Nonlinear differential equations and systems play a crucial role in modeling systems where time-dependent factors exhibit nonlinear characteristics. Due to their nonlinear nature, solving such systems often presents significant difficulties and challenges. In this study, we propose a method utilizing Physics-Informed Neural Networks (PINNs) to solve the nonlinear energy supply-demand (ESD) system. We design a neural network with four outputs, where each output approximates a function that corresponds to one of the unknown functions in the nonlinear system of differential equations describing the four-dimensional ESD problem. The neural network model is then trained and the parameters are identified, optimized to achieve a more accurate solution. The solutions obtained from the neural network for this problem are equivalent when we compare and evaluate them against the Runge-Kutta numerical method of order 4/5 (RK45). However, the method utilizing neural networks is considered a modern and promising approach, as it effectively exploits the superior computational power of advanced computer systems, especially in solving complex problems. Another advantage is that the neural network model, after being trained, can solve the nonlinear system of differential equations across a continuous domain. In other words, neural networks are not only trained to approximate the solution functions for the nonlinear ESD system but can also represent the complex dynamic relationships between the system's components. However, this approach requires significant time and computational power due to the need for model training.
\end{abstract}

% keywords can be removed

\section{Introduction} \label{sec:intro}
Differential equations and their systems are powerful mathematical methods used to model problems across various real-world fields such as physics, engineering, economics, healthcare, energy, and others \cite{ref-chicone,ref-wong,ref-zachmanoglou}. In the energy sector, the development of mathematical models for managing efficient energy supply and demand plays a crucial role in the flexible energy community with energy storage systems and renewable energy generation \cite{ref-tomin2002}. One notable application of differential equation systems is modeling the ESD system, which describes the dynamic relationship between energy supply, demand, and the distribution of energy across different regions based on development indicators of each area. The behavior of these variables is modeled through a system of differential equations, as formulated by Mei Sun et al \cite{ref-sun2007,ref-sun2005}. Research has shown that the ESD system exhibits chaotic properties \cite{ref-sun2009}, and the solutions to these differential equations are characterized by strong nonlinearity, making it highly complex to find an exact analytical solution. In practice, solving such systems of differential equations relies on approximate methods. The traditional mathematical approach in this case involves numerical methods. Most of these methods require discretizing the time and space domains into grids \cite{ref-vuik,ref-lyengar}, so the solutions provided by numerical methods are often discrete sets of values. Additionally, numerical methods can face difficulties when solving complex nonlinear systems of equations.

Recently, with the explosive and increasingly powerful development of artificial intelligence, machine learning, and particularly deep learning methods, these techniques are being applied across numerous fields of science and engineering. One of the recent promising studies involves using deep neural networks to solve differential equations and their systems. Isaac et al \cite{ref-lagaris} introduced a novel approach, utilizing neural networks to solve both ordinary differential equations (ODEs) with initial and boundary conditions and partial differential equations (PDEs). Since then, many related studies have employed neural networks and developed them into the PINNs method \cite{ref-raissi2019,ref-raissi2017} to solve complex problems such as \cite{ref-raissi2018,ref-margenberg,ref-hu,ref-farkane,ref-eivazi,ref-lu}: the Navier-Stokes equations, Burgers' equation, Schrödinger equation, Poisson equation, diffusion equation, Lorenz equation, and Volterra equations. Apart from dynamic models based on differential and differential-algebraic equations, many inverse problems can be formulated in terms of integral equations \cite{ref-sidorov}. Research on applying neural networks to solve integro-differential equations has also been conducted \cite{ref-yuan,ref-li}.

The advantage of this method over traditional numerical approaches is its ability to solve complex problems, especially systems of strongly nonlinear equations, without the need to discretize the time and space domains \cite{ref-matthews,ref-baty}. Initially, the neural network only needs to be trained to solve the system of equations at a set of specific time points in the computational domain. Once the model is built, it can provide solutions for any point within the trained time domain, that means the solution of the neural network method is a continuous domain \cite{ref-raissi2019,ref-uriarte,ref-gorikhovskii}. In other words, the neural network can be trained to approximate the solution function of the system of differential equations. Moreover, the flexibility of PINNs allows for training a model that integrates constraints from the system of differential equations with real-world data \cite{ref-matthews,ref-baty,ref-uriarte,ref-gorikhovskii}, which is particularly useful for modeling real-world applications such as the energy supply-demand problem. This helps to predict the supply-demand variables of the system in a way that is closer to reality. Given these considerations, applying PINNs to the energy supply-demand problem represents a new and highly promising direction.

In this study, we employ a deep artificial neural network based on the concept of PINNs to solve the nonlinear ESD system of differential equations. The structure of this paper consists of five main parts: The Section \ref{sec:intro} provides an overview of the research, the Section \ref{sec:problem} presents the problem to be solved, which is a system of differential equations describing the ESD system. In the Section \ref{sec:methods}, we introduce the deep learning method based on neural networks and PINNs to solve the system of differential equations. In the Section \ref{sec:model}, we delve into the construction of a deep learning model to address the nonlinear ESD system. This involves designing the neural network architecture, defining an appropriate loss function, and training the model. The Section \ref{sec:results} presents the results and evaluations.
%%%%%%%%%%%%%%%%
\section{Problem description}\label{sec:problem}
The four-dimensional ESD system is a complex model that describes the time-varying relationship between energy supply and demand, as well as the distribution of energy between different regions, E and F. This dynamic relationship is represented by the following system of differential equations \cite{ref-sun2007}:
\begin{equation}\label{eq1}
	\begin{cases}
		{{x}_{1}}^{\prime }(t)={{a}_{1}}{{x}_{1}(t)}(1-\frac{{{x}_{1}(t)}}{M})-{{a}_{2}}({{x}_{2}(t)}+{{x}_{3}(t)})-{{d}_{3}}{{x}_{4}(t)} \\
		{{x}_{2}}^{\prime }(t)=-{{z}_{1}}{{x}_{2}(t)}-{{z}_{2}}{{x}_{3}(t)}+{{z}_{3}}{{x}_{1}(t)}[N-({{x}_{1}(t)}-{{x}_{3}(t)})] \\
		{{x}_{3}}^{\prime }(t)={{s}_{1}}{{x}_{3}(t)}({{s}_{2}}{{x}_{1}(t)}-{{s}_{3}}) \\
		{{x}_{4}}^{\prime }(t)={{d}_{1}}{{x}_{1}(t)}-{{d}_{2}}{{x}_{4}(t)} \\
	\end{cases}
\end{equation}
where: ${{x}_{1}}(t)$ is a function representing the time-varying energy resource demand of region F, ${{x}_{2}}(t)$ is a function representing the time-varying energy resource supply from region E to region F, ${{x}_{3}}(t)$ is a function representing the time-varying energy resource imports of region F, and ${{x}_{4}}(t)$ is a function representing the time-varying renewable energy resources of region F. ${{a}_{i}},{{d}_{i}},{{z}_{i}},{{s}_{i}},N,M>0$ are positive constants and $N<M$. With the coefficients ${{a}_{1}}=0.09$, ${{a}_{2}}=0.15$, ${{z}_{1}}=0.06$, ${{z}_{2}}=0.082$, ${{z}_{3}}=0.07$, ${{s}_{1}}=0.2$, ${{s}_{2}}=0.5$, ${{s}_{3}}=0.4$, $M=1.8$, $N=1$, ${{d}_{1}}=0.1$, ${{d}_{2}}=0.06$, ${{d}_{3}}=0.08$ and the initial conditions ${{x}_{1}}(0)=0.82$, ${{x}_{2}}(0)=0.29$, ${{x}_{3}}(0)=0.48$, ${{x}_{4}}(0)=0.1$. The system \eqref{eq1} is in a chaotic state \cite{ref-sun2007,ref-sun2009}. 

Our objective is to determine the time series values described by the energy supply-demand problem by solving the system of differential equations \eqref{eq1} under chaotic conditions with the provided parameters.
%%%%%%%%%%%%%%%%
\section{Methods}\label{sec:methods}
\subsection{Deep learning neural networks:}
A deep neural network is characterized by an architecture consisting of multiple interconnected layers of neurons, where the connections between these layers are represented by a set of weights. A typical architecture of a deep neural network generally includes three main layers \cite{ref-goodfellow}: The input layer, which is the first layer that receives the data. In most problems, the input data can be features of the objects or the data to be computed. In this case, the input data consists of different time value points that need to be calculated. The hidden layer performs complex computations, including nonlinear operations through activation functions \cite{ref-chollet}, to learn and extract features from the data. The output layer contains the results to be predicted or calculated; In this problem, these are the solutions that need to be found for the system of differential equations describing the four-dimensional energy supply-demand system. During the training phase, the neural network's weights are gradually adjusted through the training iterations to minimize the loss function \cite{ref-geron}, which quantifies the error between the network's output and the constraints from the system of differential equations or the desired output values. The neural network accomplishes this through two processes known as forward propagation \cite{ref-nielsen}, where data is transmitted from the input layer through the hidden layers and finally to the output layer, the second process called backward propagation \cite{ref-nguyen}, which occurs when the network has computed the errors of the loss function. The network will make adjustments and update the weights to minimize the error value by calculating the Gradient Descent and using optimization algorithms such as \cite{ref-goodfellow,ref-chollet,ref-geron,ref-nielsen,ref-nguyen}: Stochastic Gradient Descent (SGD), RMSprop, Adam, or LBFGS..etc.

\subsubsection{The generalized model of a neural network:}
Consider a neural network where each hidden layer is denoted as $L$ , the $v$-th ($v\ge 1;v\in \mathbb{Z}$) hidden layer is denoted as ${{L}^{(v)}}$. Let the number of nodes in the $v$-th hidden layer be denoted as ${{n}^{(v)}}$. 

Let the weight matrix between layer ${{L}^{(v-1)}}$ and layer ${{L}^{(v)}}$ be denoted as ${{W}^{(v)}}$ ( the matrix ${{W}^{(v)}}$ will have dimensions ${{n}^{(v-1)}}\times {{n}^{(v)}}$ ), where each element $W_{ij}^{(v)}$ of the weight matrix represents a connection from the $i$-th node ($1\le i\le {{n}^{(v-1)}}$) of layer ${{L}^{(v-1)}}$ to the $j$-th node ($1\le j\le {{n}^{(v)}}$) of layer ${{L}^{(v)}}$. 

Let ${{b}^{(v)}}$ be a one-dimensional vector with ${{n}^{(v)}}$ elements, where the vector ${{b}^{(v)}}$ includes a set of bias values for each node in layer ${{L}^{(v)}}$

Each node in the neural network is designed to perform two calculations as follows \cite{ref-baty,ref-geron,ref-nguyen}:

\vspace{0.2cm}
\textit{* The first operation:} 
\begin{equation}\label{eq2}
	z_{j}^{(v)}=\sum\limits_{i=1}^{{{n}^{(v-1)}}}{a_{i}^{(v-1)}}\times W_{ij}^{(v)}+b_{j}^{(v)} ,
\end{equation}

where: $z_{j}^{(v)}$ is the linear sum of the product of the output values of the nodes in layer ${{L}^{v-1}}$ and their corresponding weights to the $j$-th node of layer ${{L}^{(v)}}$, plus the bias term of the node being considered.  

$W_{ij}^{(v)}$: is the weight connecting the $i$-th node of layer ${{L}^{(v-1)}}$ to the $j$-th node of layer ${{L}^{(v)}}$.

$b_{j}^{(v)}$: is the bias value of the $j$-th node in ${{L}^{(v)}}$.

\vspace{0.2cm}
\textit{* The second operation uses nonlinear activation functions:}
\begin{equation}\label{eq3}
	a_{j}^{(v)}=\sigma (z_{j}^{(v)}) \:,
\end{equation}

where: $a_{j}^{(v)}$ is the output value of the $j$-th node in layer ${{L}^{(v)}}$

\hspace*{2.9em} $\sigma $: is an activation function 

In neural networks, activation functions $\sigma $ play an important role as they help represent nonlinear relationships between the nodes of the neural network \cite{ref-goodfellow,ref-nielsen}. The Sigmoid, Tanh, ReLU, and Softmax functions \cite{ref-chollet,ref-geron,ref-nielsen,ref-nguyen} are a few examples of frequently used nonlinear activation functions. Each function is appropriate for the particular needs of various issues. In the experimental part of this study, we used the tanh (Hyperbolic Tangent) activation function. The following is the formula for the tanh function, which transforms the input variables into nonlinear values within the range (-1, 1) \cite{ref-nielsen}:
\begin{equation}\label{eq4}
	\tanh (x)=\frac{{{e}^{x}}-{{e}^{-x}}}{{{e}^{x}}+{{e}^{-x}}} \:.
\end{equation}

\subsubsection{The process of optimizing the parameters of a neural network:}

The updating of weights in the neural network is performed through the backpropagation process, which consists of two main steps. In the first step, neural networks compute the partial derivatives of the loss function with respect to each weight within the network. This operation is computed backward from the output layer to the input layer using the chain rule \cite{ref-goodfellow,ref-chollet}. Considering a specific layer:
\begin{equation}\label{eq5}
	\frac{\partial L}{\partial {{w}_{ij}}}=\frac{\partial L}{\partial {{a}_{j}}}\cdot \frac{\partial {{a}_{j}}}{\partial {{z}_{j}}}\cdot \frac{\partial {{z}_{j}}}{\partial {{w}_{ij}}} \:,
\end{equation}

\noindent where: 

\hspace*{2.0em}$\frac{\partial L}{\partial {{w}_{ij}}}$ is the partial derivative of the loss function with respect to the $i$-th weight of neuron $j$. 

\hspace*{2.0em}$\frac{\partial L}{\partial {{a}_{j}}}$ is the partial derivative of the loss function with respect to the output (according to the activation function) at the node with the weight being considered. 

\hspace*{2.0em}$\frac{\partial {{a}_{j}}}{\partial {{z}_{j}}}$ is the derivative of the output value at the $j$-th node with respect to the sum function ${{z}_{j}}$. 

\hspace*{2.0em}$\frac{\partial {{z}_{j}}}{\partial {{w}_{ij}}}$ is the derivative of the sum ${{z}_{j}}$ at the $j$-th node with respect to the weight being considered.

%\vspace{0.2em}
The second step in updating the weights of the neural network is based on the Gradient Descent optimization algorithm. The weights of the network will be updated according to the following formula \cite{ref-goodfellow,ref-nielsen,ref-nguyen}:
\begin{equation}\label{eq6}
	{{w}_{update}}={{w}_{old}}-\eta \cdot \frac{\partial L}{\partial w} \:,
\end{equation}

\noindent where: $w$ is the weight to be determined in the network,

\vspace{0.1em}
\hspace*{2.7em} $\eta $ is learning rate,

\vspace{0.1em}
\hspace*{2.7em} $\frac{\partial L}{\partial w}$ is the partial derivative of the loss function concerning the weight.

\vspace{0.2em}
The updating of bias values is performed in the same manner as for weights. The process of updating the weights and biases of the neural network is repeated multiple times across each epoch until the loss function decreases to a desired value or until convergence is achieved \cite{ref-goodfellow,ref-nielsen}.

\subsection{Physics-Informed Neural Networks (PINNs)}

PINNs are a unique class of neural networks designed by combining traditional neural networks and physical or mathematical models. Its main idea is that physical constraints and conditions are directly integrated into the neural network through the loss function \cite{ref-matthews,ref-baty,ref-uriarte,ref-gorikhovskii}. To handle and compute complex derivative operations integrated into the loss function of PINNs, we use Automatic Differentiation \cite{ref-hu,ref-lu,ref-yuan}, which is a powerful technique that allows for the computation of high-order and complex derivatives.

The process of building and training a PINNs to solve a system of differential equations is carried out through the following main steps:

\textit{Constructing and designing the neural network:} Construct a neural network with input and output layers adapted to each specific problem to be solved, in which each unknown function in the system of differential equations is approximated by the respective outputs of the neural network. The architecture, including the number of hidden layers, the nodes per layer, and their activation functions, is defined and optimized to obtain solutions that satisfy the system of equations.

\textit{Determining physical constraints:} The differential equation system, initial conditions, and boundary conditions are used to integrate into the neural network's loss function as constraints.

\textit{Optimizing the loss function:} Select and use appropriate optimization algorithms to minimize the loss function.

\textit{Predicting the solution:} A trained neural network can be used to compute and predict the solution of the differential equation system.
%%%%%%%%%%%%%%%%%%%%%%
\section{Model Building}\label{sec:model} 

In this study, we propose a solution based on the concept of PINNs, designing a deep neural network to solve the nonlinear system of differential equations that describes the ESD system \eqref{eq1} through four main steps:

\textbf{Step 1}: A neural network for this problem is designed with 4 outputs. In this study, we describe it as a mathematical function $NN(t,Wb)$, this function, or in other words, this neural network, is dependent on two variables: the time variable $t$ represents the input data of the neural network, and $Wb$ represents the set of weights, biases, and parameters that need to be determined for the neural network. Each unknown solution function of the nonlinear differential equation system \eqref{eq1} describing the 4-dimensional ESD system is approximated by a corresponding output of the network through the training process. These neural network outputs constitute a vector function described as follows: 
\[OutPut(NN(t,Wb))=[{{X}_{1}}(t,Wb),{{X}_{2}}(t,Wb),{{X}_{3}}(t,Wb),{{X}_{4}}(t,Wb)]\]

The objective is to find the parameters $Wb$ such that the values of the functions generated by the neural network satisfy the following condition:

\begin{center}
	${{X}_{1}}(t,Wb)\approx {{x}_{1}}(t)$, ${{X}_{2}}(t,Wb)\approx {{x}_{2}}(t)$, ${{X}_{3}}(t,Wb)\approx {{x}_{3}}(t)$, ${{X}_{4}}(t,Wb)\approx {{x}_{4}}(t)$.
\end{center}

\textbf{Step 2}: Define the time domain for computation as $t\in [a,b]$, divide this domain into $N$ consecutive points with different values of $t$ (${{t}_{0}}<{{t}_{1}}<{{t}_{2}}<...<{{t}_{N-1}}$ where ${{t}_{0}}=a$, ${{t}_{N-1}}=b$). These values are subsequently input into the neural network for training and calculation.

\textbf{Step 3}: Determine the constraints and design the loss function

The constraints satisfying the mathematical conditions of the nonlinear differential equation system \eqref{eq1} and the initial conditions are incorporated into the loss function.

\vspace{0.3em}
\textbf{\textit{a) The constraints are integrated into the loss function to satisfy the differential equation system defined as follows:}} 
\begin{center}
	\resizebox{0.99\textwidth}{!}{
		$Loss_{X1\_eq}=\frac{1}{N}{{\sum\limits_{i=1}^{N}{\left\| {{X}_{1}}^{\prime }({{t}_{i}},Wb)-[{{a}_{1}}{{X}_{1}}({{t}_{i}},Wb)(1-\frac{{{X}_{1}}({{t}_{i}},Wb)}{M})-{{a}_{2}}({{X}_{2}}({{t}_{i}},Wb)+{{X}_{3}}({{t}_{i}},Wb))-{{d}_{3}}{{X}_{4}}({{t}_{i}},Wb)] \right\|}}^{2}}$ 
	}
\end{center}
\begin{center}
	\resizebox{0.99\textwidth}{!}{
		$Loss_{X2\_eq}=\frac{1}{N}{{\sum\limits_{i=1}^{N}{\left\| {{X}_{2}}^{\prime }({{t}_{i}},Wb)-(-{{z}_{1}}{{X}_{2}}({{t}_{i}},Wb)-{{z}_{2}}{{X}_{3}}({{t}_{i}},Wb)+{{z}_{3}}{{X}_{1}}({{t}_{i}},Wb)[N-({{X}_{1}}({{t}_{i}},Wb)-{{X}_{3}}({{t}_{i}},Wb))]) \right\|}}^{2}}$
	}
\end{center}
\begin{center}
	\resizebox{0.67\textwidth}{!}{
		$Loss_{X3\_eq}=\frac{1}{N}{{\sum\limits_{i=1}^{N}{\left\| {{X}_{3}}^{\prime }({{t}_{i}},Wb)-[{{s}_{1}}{{X}_{3}}({{t}_{i}},Wb)({{s}_{2}}{{X}_{1}}({{t}_{i}},Wb)-{{s}_{3}})] \right\|}}^{2}}$
	}
\end{center}
\begin{center}
	\resizebox{0.67\textwidth}{!}{
		$Loss_{X4\_eq}=\frac{1}{N}{{\sum\limits_{i=1}^{N}{\left\| {{X}_{4}}^{\prime }({{t}_{i}},Wb)-[{{d}_{1}}{{X}_{1}}({{t}_{i}},Wb)-{{d}_{2}}{{X}_{4}}({{t}_{i}},Wb)] \right\|}}^{2}}$
	}
\end{center}

where $Loss_{X1\_eq}$, $Loss_{X2\_eq}$, $Loss_{X3\_eq}$, and $Loss_{X4\_eq}$ are constraint functions that measure the error for the four output values of the neural network, satisfying the mathematical conditions of the nonlinear differential equation system \eqref{eq1}. The objective is to ensure that, when the solutions generated by the neural network are substituted into the system of equations \eqref{eq1}, the Mean Squared Error (MSE) \cite{ref-goodfellow} between the left and right sides of the equations is minimized as much as possible.

\vspace{0.3em}
\textbf{\textit{b) The constraints are integrated into the loss function to satisfy the initial conditions defined as follows:}}
\[Loss_{initial}=\sum\limits_{i=1}^{4}{{{\left\| {{X}_{i}}({{t}_{initial}},Wb)-{{x}_{i}}({{t}_{initial}}) \right\|}^{2}}}\]

where: $Loss_{initial}$ is a constraint function measuring the error between the neural network’s output values and initial conditions of the system. This function is computed based on the squared error between the solutions generated by the neural network when the variable $t$ is at the initial time point and the given initial condition values of the functions to be determined in the system of equations. The objective is also to ensure that this error value is minimized.

\vspace{0.2em}
${{t}_{initial}}$: is the value of the time variable at the initial time point.

\vspace{0.2em}
${{X}_{i}}({{t}_{initial}},Wb)$: is the value of the $i$-th output of neural network when the variable $t$ is at the initial time point.

\vspace{0.2em}
${{x}_{i}}({{t}_{initial}})$: is the value of the $i$-th nonlinear functions in the system of equations \eqref{eq1} when the variable $t$ is at the initial time point.  

\vspace{0.3em}
\textbf{\textit{c) Define the total loss function for the neural network:}}
\[Los{{s}_{total}}=\alpha (Los{{s}_{X1\_eq}}+Los{{s}_{X2\_eq}}+Los{{s}_{X3\_eq}}+Los{{s}_{X4\_eq}})+\beta Los{{s}_{initial}}\]
where: $\alpha $ and $\beta $ are real-valued parameters, selecting and adjusting these parameters appropriately will enable the model to focus on higher-priority conditions, thereby improving convergence speed and accuracy. 

\begin{figure}[h] 
	\centering
	\includegraphics[width=0.90\textwidth]{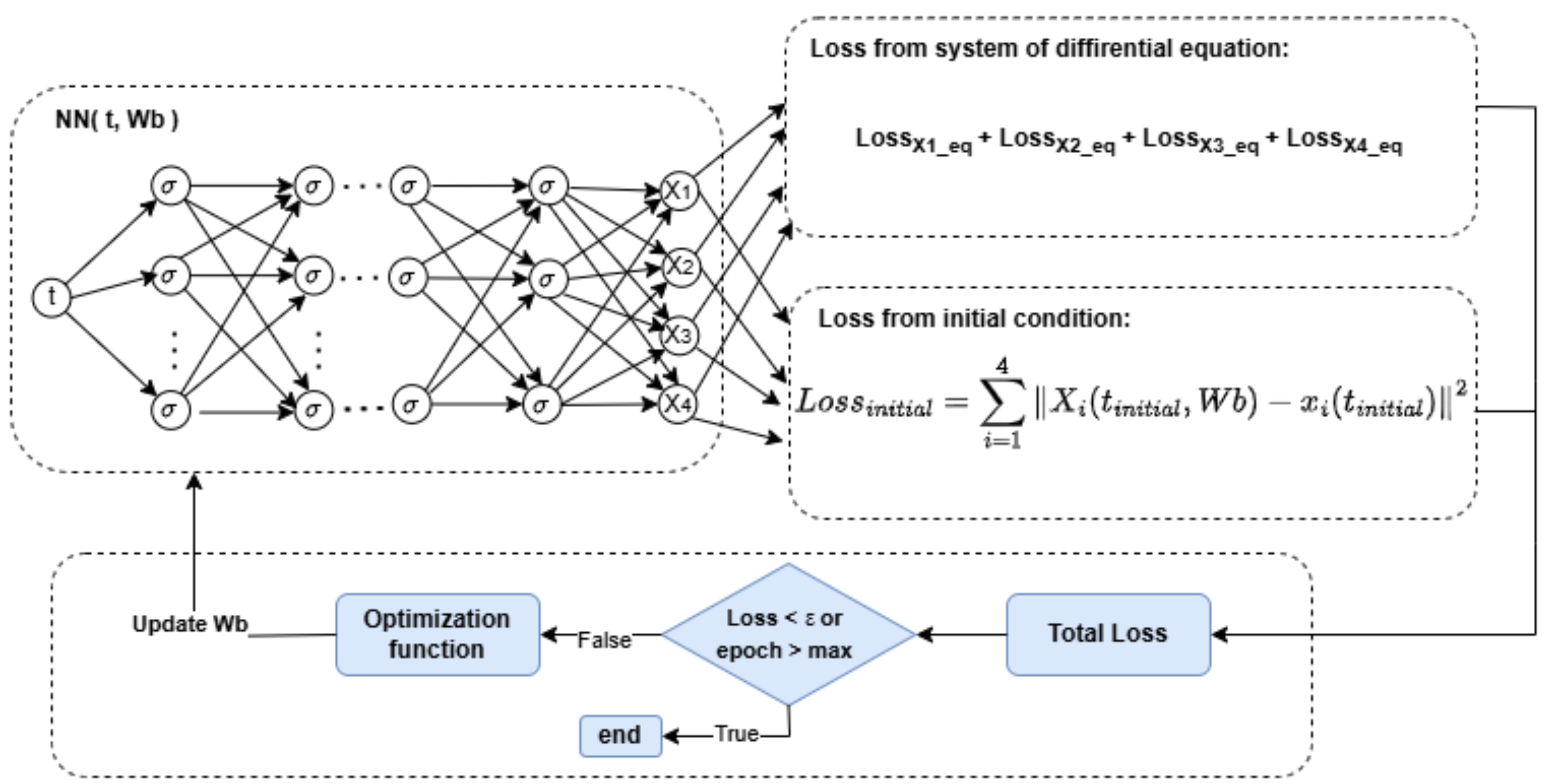} 
	\captionsetup{justification=centering}
	\caption{Overview of the method, where $\varepsilon$ is the desired value to be achieved when minimizing the loss function, and $max$ is the maximum limit of the number of training epochs}
	\label{Fig-01}
\end{figure}

\vspace{0.2em}
\textbf{Step 4}: Optimization algorithms are used to train the deep learning network to find the best parameters of the model in order to minimize the total loss function. In this study, we utilize the Adam optimization algorithm integrated within the TensorFlow library \cite{ref-chollet,ref-geron}. Figure \ref{Fig-01} shows an overview of the method.

%%%%%%%%%%%%%%%%%%%%%%
\section{Results and Evaluation}\label{sec:results} 

\begin{figure}[h] 
	\centering
	\includegraphics[width=0.70\textwidth]{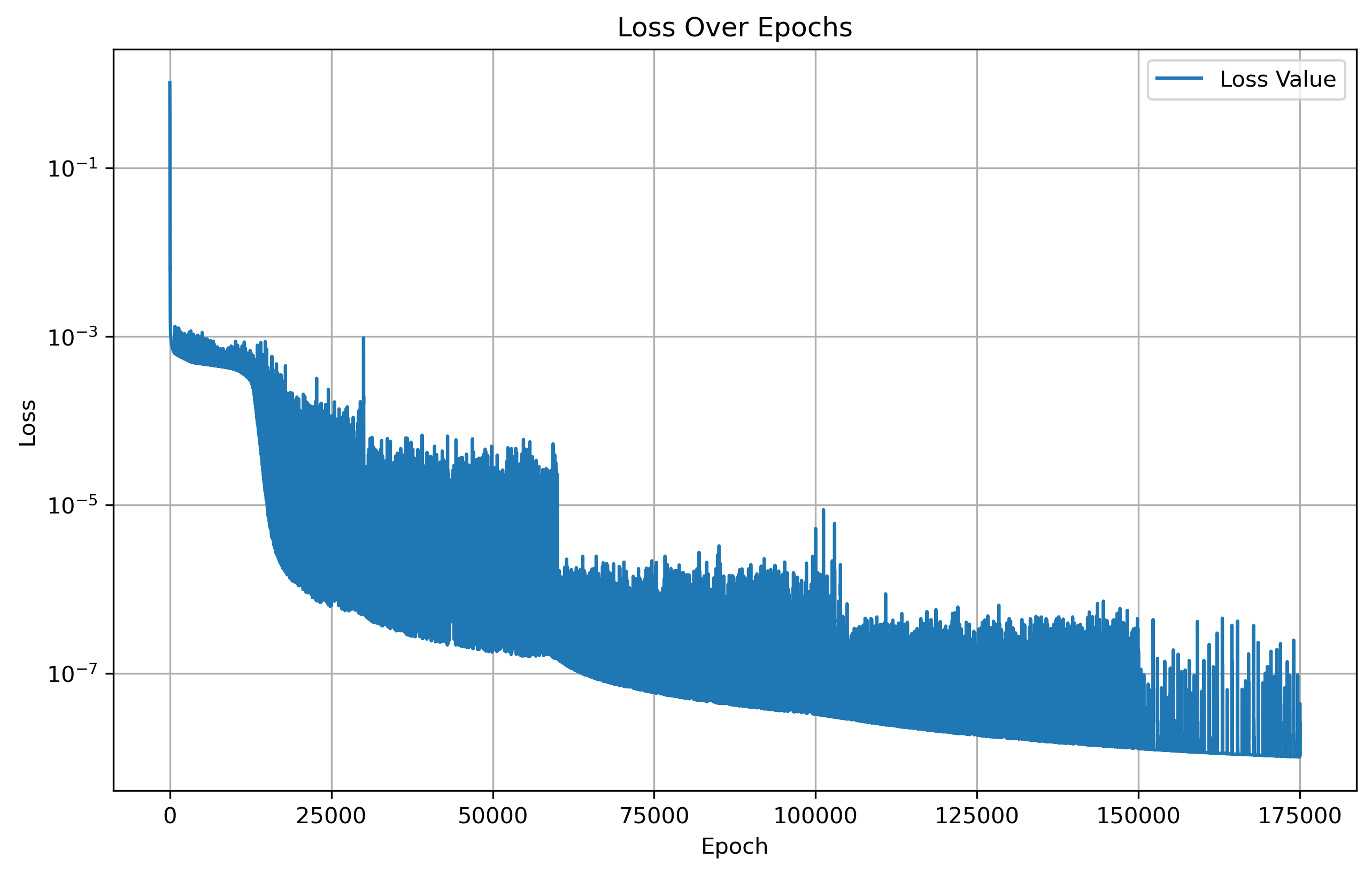} 
	\captionsetup{justification=centering}
	\caption{A chart describing the value of the loss function over the training epochs}
	\label{Fig-02}
\end{figure}

\begin{figure}[h] 
	\centering
	\includegraphics[width=0.65\textwidth]{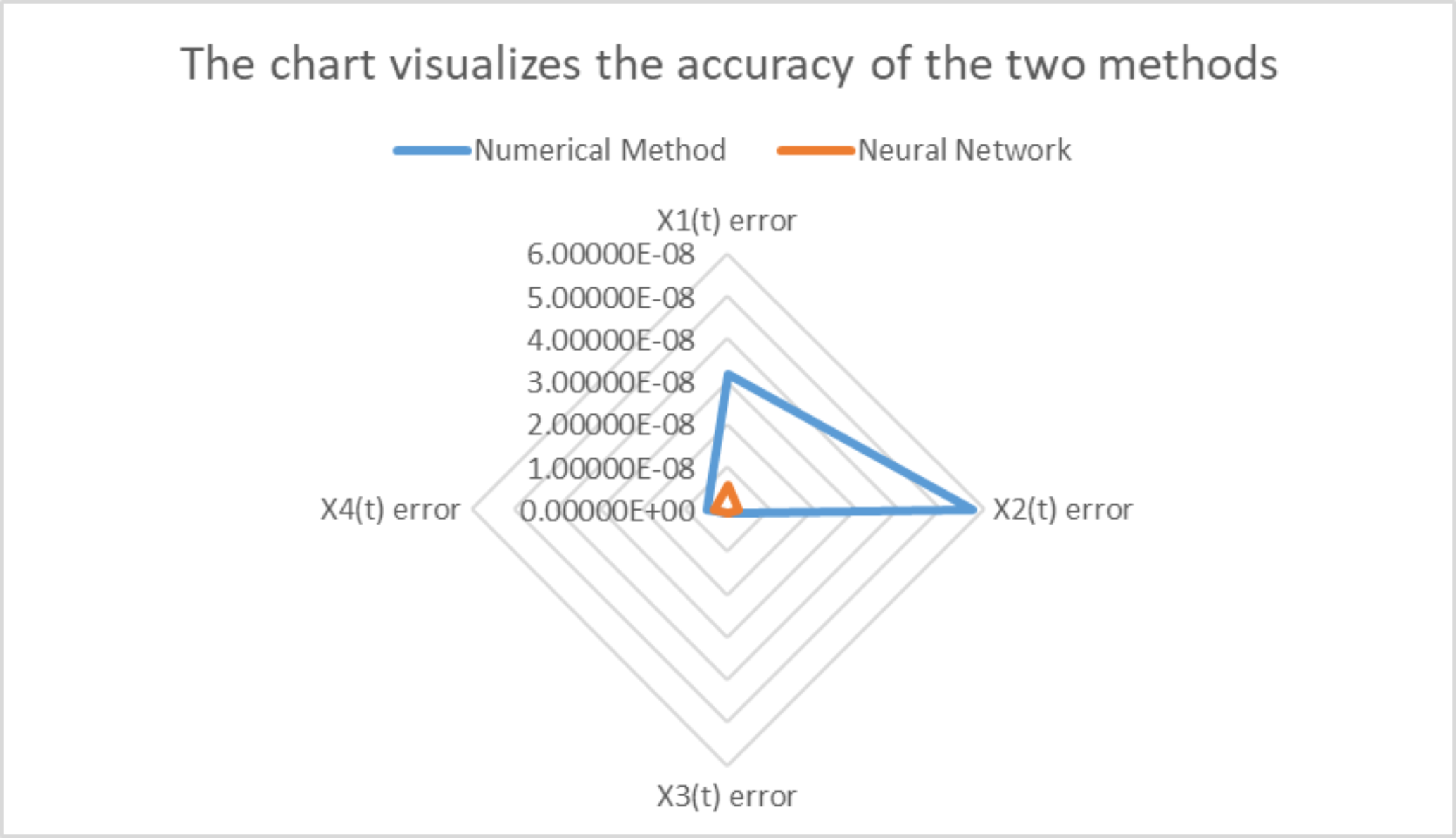} 
	\captionsetup{justification=centering}
	\caption{The chart visualizes a comparison of the accuracy of the two methods}
	\label{Fig-03}
\end{figure}

In this study, we conducted an experiment to construct a neural network with an architecture consisting of an input layer that receives different time data points, 16 hidden layers with 100 neurons each, and an output layer with 4 neurons. The neurons in the output layer represent the time-dependent values of the four functions to be determined in the system of equations \eqref{eq1}. We set $\alpha =10$, $\beta =1$ and used the Adam optimizer \cite{ref-geron,ref-nielsen} to minimize the loss function. The time interval $t = [0,100]$ was divided into $N = 20,000$ equally spaced time data points. For the numerical method, in this study, we use the SciPy library \cite{ref-SciPy} to employ the RK45 method \cite{ref-lyengar,ref-shampine}, which combines the fourth-order and fifth-order Runge-Kutta formulas to achieve high accuracy and efficiency. This method allows for the adaptive adjustment of step size to meet the required accuracy while optimizing computational time. We solved system \eqref{eq1} using the RK45 numerical method with an absolute tolerance of $1\times {{10}^{-06}}$ and a relative tolerance of $1\times {{10}^{-03}}$, over the same time domain with $N$ time points, as used by the neural network method. The solutions from both methods were then compared and evaluated. When using the neural network method, we applied a learning rate schedule, where the initial learning rate was set to $8\times {{10}^{-05}}$, and gradually reduced over the training epochs, with the minimum learning rate being $1\times {{10}^{-06}}$. After $175,000$ epochs, Figure \ref{Fig-02} shows the value of the loss function over the training epochs. We observed that the neural network method, with a simple network architecture in our experiment, began to provide more accurate solutions than the RK45 method for all four solutions sought, as presented in Table \ref{tab1} and illustrated in Figure \ref{Fig-03}.

\clearpage

To conduct a comparative analysis of the error between two methods, we apply the finite difference method, supported by the NumPy library \cite{ref-chollet,ref-geron}, to approximate the derivatives of these solutions with respect to the time variable $t$. Then, we substitute the solutions obtained from both methods into the original system of equations \eqref{eq1}. The MSE method is used to calculate the difference between the left-hand side and the right-hand side. The smaller the MSE value, the smaller the error between the two sides, indicating that the obtained solution better satisfies the original system of equations.

\begin{table}[h] 
	\renewcommand{\arraystretch}{1.4}
	\caption{Comparing the accuracy of the results between the solutions obtained using the numerical method and those obtained using the neural network method. \label{tab1}}
	\resizebox{1.0\textwidth}{!}{
	\begin{tabular}{ccccc}
		\toprule
		\textbf{Method}	& \textbf{$X1(t)$ error}	& \textbf{$X2(t)$ error} & \textbf{$X3(t)$ error} & \textbf{$X4(t)$ error} \\
		\midrule
		Numerical method & $3.16804\times {{10}^{-08}}$ & $5.73393\times {{10}^{-08}}$ & $7.70381\times {{10}^{-10}}$ & $4.67541\times {{10}^{-09}}$ \\ 
		Neural network & $5.67513\times {{10}^{-09}}$ & $2.43858\times {{10}^{-09}}$ & $7.54885\times {{10}^{-10}}$ & $3.10772\times {{10}^{-09}}$ \\
		\bottomrule
	\end{tabular}
	}
\end{table}

Additionally, the direct comparison of the errors between the solutions obtained from the two methods is performed and evaluated using the following metrics \cite{ref-wang}: R-squared, MAE (Mean Absolute Error), MSE (Mean Squared Error), and RMSE (Root Mean Squared Error):

\[{{R}^{2}}=1-\frac{\sum\nolimits_{i=1}^{N}{{{({{y}_{i}}-{{{\hat{y}}}_{i}})}^{2}}}}{\sum\nolimits_{i=1}^{N}{{{({{y}_{i}}-\bar{y})}^{2}}}}\]

\[MAE=\frac{1}{N}\sum\limits_{i=1}^{N}{|{{y}_{i}}-{{{\hat{y}}}_{i}}}|\]

\[MSE=\frac{1}{N}\sum\limits_{i=1}^{N}{({{y}_{i}}}-{{\hat{y}}_{i}}{{)}^{2}}\]

\[RMSE=\sqrt{\frac{1}{N}\sum\limits_{i=1}^{N}{{{({{y}_{i}}-{{{\hat{y}}}_{i}})}^{2}}}}\]

where: ${{y}_{i}}$ is the solution value obtained using the numerical method.

\vspace{0.2em}
\hspace*{2.9em} ${{\hat{y}}_{i}}$ is the solution value obtained using the neural network method.

\vspace{0.2em}
\begin{minipage}{\textwidth}
	\setlength{\leftskip}{3.3em}
	\setlength{\rightskip}{1.5em}
	\noindent $\bar{y}$ is the average value of all the solution values obtained using the numerical method.
\end{minipage}

\vspace{0.3em}
\hspace*{2.9em} $N$ is the number of time data points

\begin{table}[h] 
	\renewcommand{\arraystretch}{1.4}
	\caption{A direct comparison between the solutions obtained using the numerical method and those obtained using the neural network method. \label{tab2}}
	\resizebox{1.0\textwidth}{!}{
		\begin{tabular}{ccccc}
			\toprule
			\textbf{Evaluation metric}	& \textbf{$X1(t)$}	& \textbf{$X2(t)$} & \textbf{$X3(t)$} & \textbf{$X4(t)$} \\
			\midrule
			R-squared & $0.99999843520$ & $0.99999507773$ & $0.99999921467$ & $0.99999709511$ \\
			MAE	  & $0.00065521649$ & $0.00064897482$ & $9.55051342325e-05$ & $0.00093267431$ \\
			MSE & $7.88852598020e-07$ & $6.61855514079e-07$ & $1.26016840498e-08$ & $1.17917245907e-06$ \\
			RMSE & $0.00088817374$ & $0.00081354503$ & $0.00011225722$ & $0.00108589708$ \\
			\bottomrule
		\end{tabular}
	}
\end{table}

\begin{figure}[h] 
	\centering
	\includegraphics[width=0.70\textwidth]{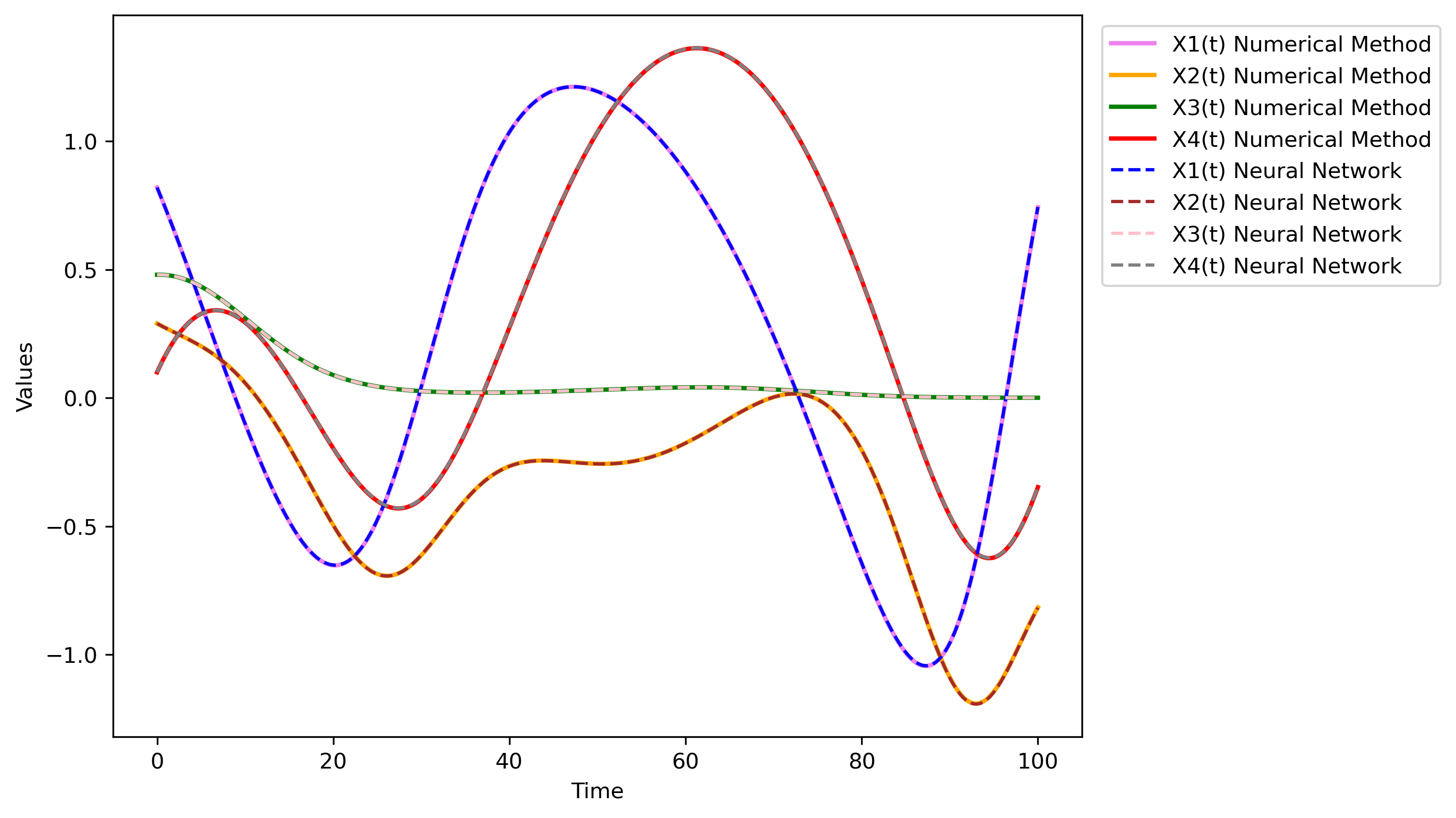} 
	\captionsetup{justification=centering}
	\caption{A general graph illustrating the direct comparison results between the RK45 numerical method and the neural network method}
	\label{Fig-04}
\end{figure}

\begin{figure}[h] 
	\centering
	\includegraphics[width=0.8\textwidth]{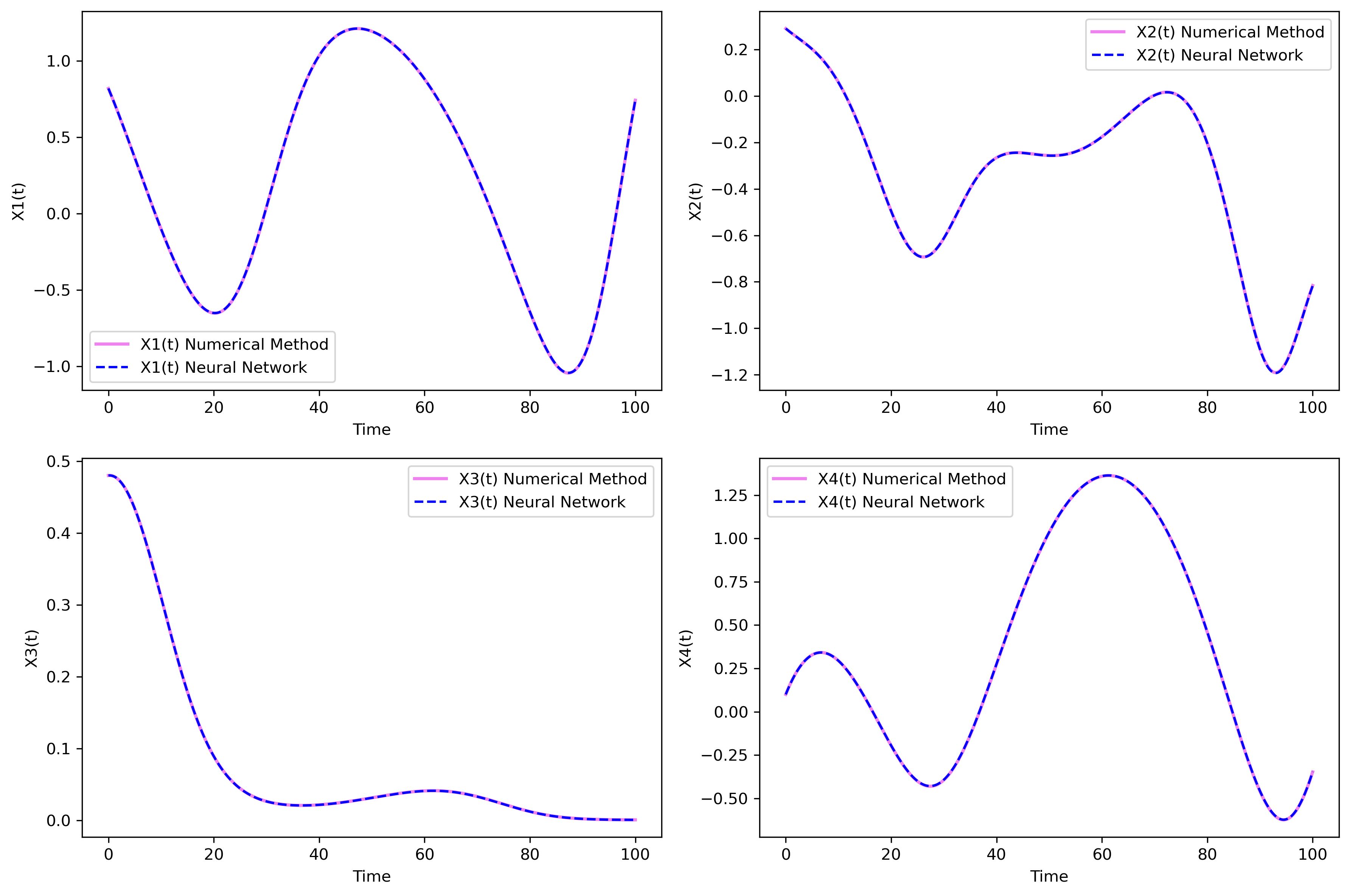} 
	\captionsetup{justification=centering}
	\caption{A detailed graph illustrating the direct comparison results between the RK45 numerical method and the neural network method}
	\label{Fig-05}
\end{figure}

The comparison results between the methods, including the RK45 numerical method and the PINNs method, presented in Table \ref{tab2} and visualized in Figures \ref{Fig-04} and \ref{Fig-05}, demonstrate that the solutions obtained from these methods are equivalent.
%%%%%%%%%%%%%%%%%%%%%%%%
\section{Conclusion}\label{sec:conclu}

In this study, we proposed a method using PINNs to solve a system of nonlinear differential equations describing the ESD system. Experimental results indicate that the PINNs method is a novel and effective approach. PINNs, a type of neural networks, present several distinct advantages, such as providing solutions over a continuous domain and the ability to leverage the computational power of modern computers. The experimental results in this study show that PINNs achieve solutions comparable to the RK45 method. Furthermore, this approach demonstrates outstanding potential. According to general methods for improving and developing deep learning models, the performance of PINNs can be enhanced by adjusting the network architecture to be more complex (such as increasing the number of hidden layers and neurons in each layer), selecting or developing suitable optimization functions, as well as increasing the data and training time. These factors will help the model learn more complex representations, playing a crucial role in fully harnessing the power of PINNs. However, this comes at the cost of greater computational time and the need for a sufficiently powerful computing system. Moreover, ensuring the stability and convergence speed of the model is a significant challenge. Overall, although this is a promising approach with high applicability in solving the nonlinear ESD system, there remain significant challenges,  particularly in enhancing model stability, optimizing convergence speed, and reducing computational power requirements. Therefore, this research topic warrants further investigation in the future.
%%%%%%%%%%%%%%%%%%%%%%%%

\section{Acknoledgement}\label{sec:fu}

The research was carried out within the state assignment of the Ministry of Science and Higher Education of the Russian Federation (project code: FZZS-2024-0003).

\bibliographystyle{unsrt}

\begin{thebibliography}{999}
	% Reference 1
	\bibitem[Chicone(1999)]{ref-chicone}
	Chicone, C. \textit{Ordinary Differential Equations with Applications}; Springer: New York, USA, 1999.
	% Reference 2
	\bibitem[Wong(2023)]{ref-wong}
	Wong, P.J.Y. \textit{Applications of Partial Differential Equations};  MDPI: Mathematics, Basel, Switzerland, 2023. [\href{https://doi.org/10.3390/books978-3-0365-9565-8}{CrossRef}]
	% Reference 3
	\bibitem[Zachmanoglou and Thoe(1986)]{ref-zachmanoglou}
	Zachmanoglou, E.C.; Thoe, D.W. \textit{Introduction to Partial Differential Equations with Applications}; Dover Publications, Inc.: New York, USA, 1986.
	% Reference 4
	\bibitem[Tomin et al.(2022)]{ref-tomin2002}
	Tomin, N.; Shakirov, V.; Kurbatsky, V.; Muzychuk, R.; Popova, E.; Sidorov, D.; Kozlov, A.; Yang, D. A multi-criteria approach to designing and managing a renewable energy community. {\em Renewable Energy} {\bf 2022}, {\em 199}, 1153--1175. [\href{https://doi.org/10.1016/j.renene.2022.08.151}{CrossRef}]
	% Reference 5
	\bibitem[Sun et al.(2007)]{ref-sun2007}
	Sun, M.; Jia, Q.; Tian, L. A new four-dimensional energy resources system and its linear feedback control. \textit{Chaos Solitons Fractals} \textbf{2007}, \textit{39}, 101--108. \href{https://www.sciencedirect.com/science/article/pii/S0960077907002585}{} [\href{https://doi.org/10.1016/j.chaos.2007.01.125}{CrossRef}]
	% Reference 6
	\bibitem[Sun et al.(2005)]{ref-sun2005}
	Sun, M.; Tian, L.; Fu, Y. An energy resources demand–supply system and its dynamical analysis. \textit{Chaos Solitons Fractals} \textbf{2005}, \textit{32}, 168--180. [\href{https://doi.org/10.1016/j.chaos.2005.10.085}{CrossRef}]
	% Reference 7
	\bibitem[Sun et al.(2009)]{ref-sun2009}
	Sun, M.; Tian, L.; Jia, Q. Adaptive control and synchronization of a four-dimensional energy resources system with unknown parameters. \textit{Chaos Solitons Fractals} \textbf{2009}, \textit{39}, 1943--1949. [\href{https://doi.org/10.1016/j.chaos.2007.06.117}{CrossRef}]
	% Reference 8
	\bibitem[Vuik et al.(2023)]{ref-vuik}
	Vuik, C.; Vermolen, F.J.; van Gijzen, M.B.; Vuik, M.J. \textit{Numerical Methods for Ordinary Differential Equations}; TUDelft: Delft University of Technology, Netherlands, 2023. [\href{https://doi.org/10.5074/t.2023.001}{CrossRef}]
	% Reference 9
	\bibitem[Lyengar and Jain(2009)]{ref-lyengar}
	Lyengar, S.R.K.; Jain, R.K. \textit{Numerical Methods}; New Age International Publishers: New Delhi, India, 2009.
	% Reference 10
	\bibitem[Lagaris et al.(1998)]{ref-lagaris}
	Lagaris, I.E.; Likas, A.; Fotiadis, D.I. Artificial Neural Networks for Solving Ordinary and Partial Differential Equations. \textit{IEEE Trans. Neural Netw.} \textbf{1998}, \textit{9}, 987--1000. [\href{https://doi.org/10.1109/72.712178}{CrossRef}]
	% Reference 11
	\bibitem[Raissi et al.(2019)]{ref-raissi2019}
	Raissi, M.; Perdikaris, P.; Karniadakis, G.E. Physics-Informed Neural Networks: A Deep Learning Framework for Solving Forward and Inverse Problems Involving Nonlinear Partial Differential Equations. \textit{J. Comput. Phys.} \textbf{2019}, \textit{378}, 686--707. [\href{https://doi.org/10.1016/j.jcp.2018.10.045}{CrossRef}]
	% Reference 12
	\bibitem[Raissi et al.(2017)]{ref-raissi2017}
	Raissi, M.; Perdikaris, P.; Karniadakis, G.E. Physics Informed Deep Learning (Part I): Data-driven Solutions of Nonlinear Partial Differential Equations. \textit{arXiv} \textbf{2017}. [\href{https://doi.org/10.48550/arXiv.1711.10561}{CrossRef}]
	% Reference 13
	\bibitem[Raissi et al.(2018)]{ref-raissi2018}
	Raissi, M.; Yazdani, A.; Karniadakis, G.E. Hidden Fluid Mechanics: A Navier-Stokes Informed Deep Learning Framework for Assimilating Flow Visualization Data. \textit{arXiv} \textbf{2018}. [\href{https://doi.org/10.48550/arXiv.1808.04327}{CrossRef}]
	% Reference 14
	\bibitem[Margenberg et al.(2022)]{ref-margenberg}
	Margenberg, N.; Hartmann, D.; Lessig, C.; Richter, T. A Neural Network Multigrid Solver for the Navier-Stokes Equations. \textit{J. Comput. Phys.} \textbf{2022}, \textit{460}. [\href{https://doi.org/10.1016/j.jcp.2022.110983}{CrossRef}]
	% Reference 15
	\bibitem[Hu and McDaniel(2023)]{ref-hu}
	Hu, B.; McDaniel, D. Applying Physics-Informed Neural Networks to Solve Navier–Stokes Equations for Laminar Flow Around a Particle. \textit{Math. Comput. Appl.} \textbf{2023}, \textit{28}. [\href{https://doi.org/10.3390/mca28050102}{CrossRef}]
	% Reference 16
	\bibitem[Farkane et al.(2023)]{ref-farkane}
	Farkane, A.; Ghogho, M.; Oudani, M.; Boutayeb, M. EPINN-NSE: Enhanced Physics-Informed Neural Networks for Solving Navier-Stokes Equations. \textit{arXiv} \textbf{2023}. [\href{https://doi.org/10.48550/arXiv.2304.03689}{CrossRef}]
	% Reference 17
	\bibitem[Eivazi et al.(2022)]{ref-eivazi}
	Eivazi, H.; Tahani, M.; Schlatter, P.; Vinuesa, R. Physics-Informed Neural Networks for Solving Reynolds-Averaged Navier–Stokes Equations. \textit{Phys. Fluids} \textbf{2022}, \textit{34}. [\href{https://doi.org/10.1063/5.0095270}{CrossRef}]
	% Reference 18
	\bibitem[Lu et al.(2021)]{ref-lu}
	Lu, L.; Meng, X.; Mao, Z.; Karniadakis, G.E. DeepXDE: A Deep Learning Library for Solving Differential Equations. \textit{SIAM Rev.} \textbf{2021}, \textit{63}, 208--228. [\href{https://doi.org/10.1137/19M1274067}{CrossRef}]
	% Reference 19
	\bibitem[Sidorov et al.(2020)]{ref-sidorov}
	Sidorov, D.; Tynda, A.; Muftahov, I.; Dreglea, A.; Liu, F. Nonlinear Systems of Volterra Equations with Piecewise Smooth Kernels: Numerical Solution and Application for Power Systems Operation. {\em Mathematics} {\bf 2020}, {\em 8}, 1257. [\href{https://doi.org/10.3390/math8081257}{CrossRef}]
	% Reference 20
	\bibitem[Yuan et al.(2022)]{ref-yuan}
	Yuan, L.; Ni, Y.-Q.; Deng, X.-Y.; Hao, S. A-PINN: Auxiliary Physics-Informed Neural Networks for Forward and Inverse Problems of Nonlinear Integro-Differential Equations. \textit{J. Comput. Phys.} \textbf{2022}, \textit{462}. [\href{https://doi.org/10.1016/j.jcp.2022.111260}{CrossRef}]
	% Reference 21
	\bibitem[Li et al.(2024)]{ref-li}
	Li, H.; Shi, P.; Li, X. Machine Learning for Nonlinear Integro-Differential Equations with Degenerate Kernel Scheme. \textit{Commun. Nonlinear Sci. Numer. Simul.} \textbf{2024}, \textit{138}. [\href{https://doi.org/10.1016/j.cnsns.2024.108242}{CrossRef}]
	% Reference 22
	\bibitem[Matthews and Bihlo(2024)]{ref-matthews}
	Matthews, J.; Bihlo, A. PinnDE: Physics-Informed Neural Networks for Solving Differential Equations. \textit{arXiv} \textbf{2024}. [\href{https://doi.org/10.48550/arXiv.2408.10011}{CrossRef}]
	% Reference 23
	\bibitem[Baty and Baty(2023)]{ref-baty}
	Baty, H.; Baty, L. Solving Differential Equations Using Physics-Informed Deep Learning: A Hands-On Tutorial with Benchmark Tests. \textit{arXiv} \textbf{2023}. [\href{https://doi.org/10.48550/arXiv.2302.12260}{CrossRef}]
	% Reference 24
	\bibitem[Uriarte(2024)]{ref-uriarte}
	Uriarte, C. Solving Partial Differential Equations Using Artificial Neural Networks. \textit{arXiv} \textbf{2024}. [\href{https://doi.org/10.48550/arXiv.2403.09001}{CrossRef}]
	% Reference 25
	\bibitem[Gorikhovskii et al.(2022)]{ref-gorikhovskii}
	Gorikhovskii, V.I.; Evdokimova, T.O.; Poletansky, V.A. Neural Networks in Solving Differential Equations. \textit{J. Phys. Conf. Ser.} \textbf{2022}, \textit{2308}, 012008. [\href{https://doi.org/10.1088/1742-6596/2308/1/012008}{CrossRef}]
	% Reference 26
	\bibitem[Goodfellow et al.(2016)]{ref-goodfellow}
	Goodfellow, I.; Bengio, Y.; Courville, A. \textit{Deep Learning}; MIT Press: Cambridge, MA, USA, 2016.
	% Reference 27
	\bibitem[Chollet(2021)]{ref-chollet}
	Chollet, F. \textit{Deep Learning with Python}, 2nd ed.; Manning Publications: Shelter Island, NY, USA, 2021.
	% Reference 28
	\bibitem[Géron(2019)]{ref-geron}
	Géron, A. \textit{Hands-On Machine Learning with Scikit-Learn, Keras, and TensorFlow}, 2nd ed.; O’Reilly Media: Sebastopol, CA, USA, 2019.
	% Reference 29
	\bibitem[Nielsen(2016)]{ref-nielsen}
	Nielsen, M. \textit{Neural Networks and Deep Learning}; 2016. Available online: \url{http://neuralnetworksanddeeplearning.com/} (accessed on 15 October 2024).
	% Reference 30
	\bibitem[Nguyen(2020)]{ref-nguyen}
	Nguyen, T.T. \textit{Basic Deep Learning}; 2020. Available online: \url{https://nttuan8.com/sach-deep-learning-co-ban/} (accessed on 01 October 2024).
	% Reference 31
	\bibitem[SciPy(2024)]{ref-SciPy}
	SciPy Reference. Available online: \url{https://docs.scipy.org/doc/scipy/reference/integrate.html} (accessed on 05 September 2024).
	% Reference 32
	\bibitem[Shampine(1986)]{ref-shampine}
	Shampine, L. W. Some Practical Runge-Kutta Formulas. {\em Mathematics of Computation} {\bf 1986}, {\em 46}, 135--150.
	% Reference 33
	\bibitem[Wang et al.(2024)]{ref-wang}
	Wang, X.; Liu, X.; Wang, Y.; Kang, X.; Geng, R.; Li, A.; Xiao, F.; Zhang, C.; Yan, D. Investigating the Deviation Between Prediction Accuracy Metrics and Control Performance Metrics in the Context of an Ice-Based Thermal Energy Storage System. \textit{J. Energy Storage} \textbf{2024}, \textit{91}. [\href{https://doi.org/10.1016/j.est.2024.112126}{CrossRef}]
\end{thebibliography}

\end{document}